\documentclass[12pt,letterpaper]{article}

\usepackage{amsfonts,amssymb}
\usepackage{authblk}
\usepackage{amsmath}
\usepackage{color}
\usepackage{url}
\usepackage{graphicx}
\usepackage[colorlinks=true,linkcolor=blue,citecolor=blue]{hyperref}
\usepackage{arydshln}

\title{Cell Detection in Microscopy Images with Deep Convolutional Neural Network and Compressed Sensing}
\author[]{Yao Xue}
\author[]{Nilanjan Ray}
\affil[]{Computing Science, University of Alberta, Canada}

\date{}

\begin{document} 

% Generate Title Page
\maketitle

\begin{abstract}

The ability to automatically detect certain types of cells or cellular subunits in microscopy images is of significant interest to a wide range of biomedical research and clinical practices. Cell detection methods have evolved from employing hand-crafted features to deep learning-based techniques. The essential idea of these methods is that their cell classifiers or detectors are trained in the pixel space, where the locations of target cells are labeled. In this paper, we seek a different route and propose a convolutional neural network (CNN)-based cell detection method that uses encoding of the output pixel space. For the cell detection problem, the output space is the sparsely labeled pixel locations indicating cell centers. We employ random projections to encode the output space to a compressed vector of fixed dimension. Then, CNN regresses this compressed vector from the input pixels. Furthermore, it is possible to stably recover sparse cell locations on the output pixel space from the predicted compressed vector using $L_1$-norm optimization. In the past, output space encoding using compressed sensing (CS) has been used in conjunction with linear and non-linear predictors. To the best of our knowledge, this is the first successful use of CNN with CS-based output space encoding. We made substantial experiments on several benchmark datasets, where the proposed CNN + CS framework (referred to as CNNCS) achieved the highest or at least top-3 performance in terms of F1-score, compared with other state-of-the-art methods.


\end{abstract}

% this dumps the abstract on a front page all by itself.
%\clearpage

%Three potential tables: Table of Contents, Table of Tables, Table of Figures.
%\tableofcontents
%\clearpage

%\listoftables
%\clearpage

%\listoffigures
%\clearpage
\section{Introduction}

Automatic cell detection is to find whether there are certain types of cells present in an input image (e.g. microscopy images) and to localize them in the image. It is of significant interest to a wide range of medical imaging tasks and clinical applications. An example is breast cancer, where the tumor proliferation speed (tumor growth) is an important biomarker indicative of breast cancer patients' prognosis. In practical scenario, the most common method is routinely performed by pathologists, who examine histological slides under a microscope based on their empirical assessments, which could be really accurate in several cases, but generally is slow and prone to fatigue induced errors.

\begin{figure}[htbp]
	\centering
	\setlength{\abovecaptionskip}{0pt}
	\setlength{\belowcaptionskip}{0pt}
	\includegraphics[width=13cm]{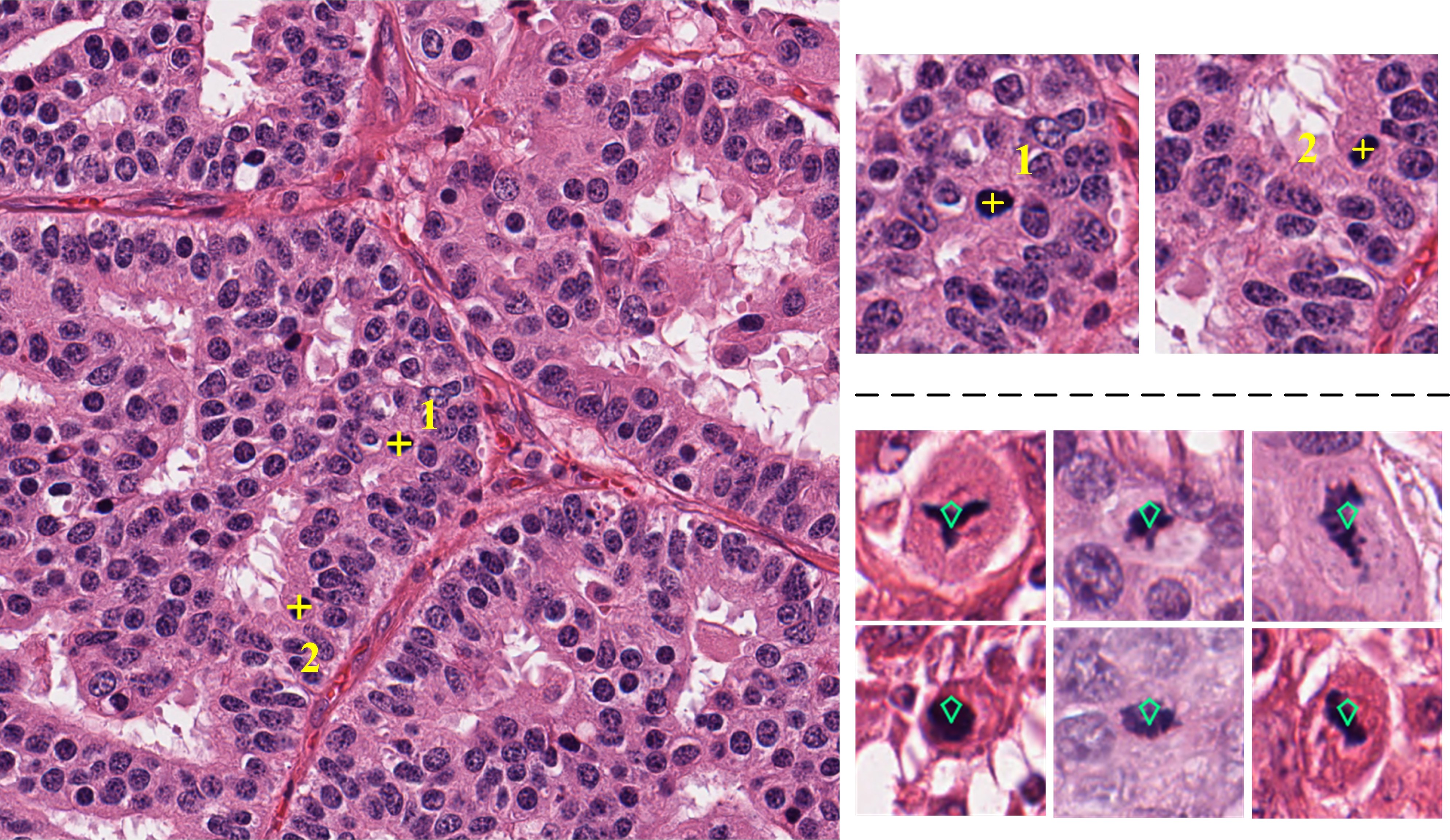}
	\caption{Left picture shows a microscopy image with two target cells annotated by yellow crosses on their centers. Right top pictures give details about the two target cells whose nuclei are in mitotic phase. Right bottom pictures provide more examples of mitotic figures.}
	\label{fig:example1}
\end{figure}

Cell detection and localization constitute several challenges that deserves our attention. First, target cells are surrounded by clutters represented by complex histological structures like capillaries,  adipocytes, collagen etc. In many cases, the size of the target cell is small, and consequently, it can be difficult to distinguish from the aforementioned clutter. Second, the target cells can appear very sparsely (only in tens), moderately densely (in tens of hundreds) or highly densely (in thousands) in a typical 2000-by-2000 pixel high resolution microscopy image as shown in Fig.~\ref{fig:example1}. Additionally, significant variations in the appearance among the targets can also be seen. These challenges render the cell detection/localization/counting problems far from being solved at the moment, in spite of significant recent progresses in computer vision research.

In recent years, object detection has been significantly advanced following the big success by deep learning. However, cell detection or localization task is not simply a sub-task of a general object detection, which typically deals with extended objects, such as humans and vehicles that occupy a significant portion of the field of view in the image. Extended object detection and localization have witnessed much progress in the computer vision community. For example, Region-based Convolutional Neural Networks (R-CNN) \cite{rcnn} and its variants \cite{fast-rcnn}, \cite{faster-rcnn}, Fully Convolutional Networks (FCN) \cite{FCN} with recent optimization \cite{redmon2016yolo9000} have become the state-of-the-art algorithms for the extended object detection problem. These solutions cannot be easily translated to cell detection, since assumptions and challenges are different for the latter. For example, for an extended object, localization is considered successful if a detection bounding box is 50\% overlapping with the actual bounding box. For cell detection, tolerance is typically on a much tighter side in order for the localization to be meaningful.

\textbf{Conventional cell detection approaches}

In the last few decades, different cell recognition methods had been proposed \cite{Meijering12}. Traditional computer vision based cell detection systems adopt classical image processing techniques, such as intensity thresholding, feature detection, morphological filtering, region accumulation, and deformable model fitting. For example, Laplacian-of-Gaussian (LoG) \cite{LoG} operator was a popular choice for blob detection; Gabor filter or LBP feature \cite{LBP} offers many interesting texture properties and had been attempted for a cell detection task \cite{LoG-cellSeg}.

Conventional cell detection approaches follow a ``hand-crafted feature representation''+``classifier'' framework. First, detection system extracts one (or multiple) kind of features as the representation of input images. Image processing techniques offer a range of feature extraction algorithms for selection. After that, machine learning based classifiers work on the feature vectors to identify or recognize regions containing target cells. ``Hand-crafted feature representation''+``classifier'' approaches suffer from the following limitations:

(1) It is a non-trivial and difficult task for humans to select suitable features. In many cases, it requires significant prior knowledge about the target cells and background.

(2) Most hand-crafted features contain many parameters that are crucial for the overall performance. Consequently, users need to perform a lot of trial-and-error experiments to tune these parameters.

(3) Usually, one particular feature is not versatile enough. The feature may often be tightly coupled with a particular type of target cell and may not work well when presented with a different type of target cell.

(4) The performance of a hand-crafted feature-based classifier soon reaches an accuracy plateau, even when trained with plenty of training data.

\textbf{Deep learning based cell detection approaches}

In comparison to the conventioal cell detection methods, deep neural networks recently has been applied to a variety of computer vision problems, and has achieved better performance on several benchmark vision datasets \cite{ImageNet}, \cite{rcnn}, \cite{FCN}. The most compelling advantage of deep learning is that it has evolved from fixed feature design strategies towards automated learning of problem-specific features directly from training data \cite{LeCun-nature}. By providing massive amount of training images and problem-specific labels, users do not have to go into the elaborate procedure for the extraction of features. Instead, deep neural network (DNN) is subsequently optimized using a mini-batch gradient descent method over the training data, so that the DNN allows autonomic learning of implicit relationships within the data. For example, shallow layers of DNN focus on learning low-level features (such as edges, lines, dots), while deep layers of DNN form more abstract high-level semantic representations (such as probability maps, or object class labels).

With the advent of deep learning in the computer vision community, it is no wonder that the state-of-the-art methods in cell detection are based on deep neural networks. The essential idea behind all these methods is that detectors are trained as a classifier in the image pixel space, either as a pixel lableing \cite{Alpher21} or as a region proposal network \cite{CasNN}. Thus, these methods predict the $\left\lbrace x,y\right\rbrace $-coordinates of cells directly on a 2-D image. Because, target cell locations are sparse in an image, the classifiers in these methods face the class imbalance issue. Moreover, target cells are often only subtly different from other cells. Thus, these methods tend produce significant amount of false positives.

\textbf{Compressed sensing-based output encoding}

In this work, deviating from past approaches, we introduce output space encoding in the cell detection and localization problem. Our observation is that the output space of cell detection is quite sparse: an automated system only needs to label a small fraction of the total pixels as cell centroid locations. To provide an example, if there are $5000$ cells present in an image of size $2000$-by-$2000$ pixels, this fraction is $5000/(2000*2000) = 0.00125$, signifying that even a dense cell image is still quite sparse in the pixel space.

Based on the observation about sparse cell locations within a microscopy image, we are motivated to apply compressed sensing (CS) \cite{Alpher16} techniques in the cell detection task. First, a fixed length, compressed vector is formed by randomly projecting the cell locations from the sparse pixel space. Next, a deep CNN is trained to predict the encoded, compressed vector directly from the input pixels (i.e., microscopy image). Then, $L_1$ norm optimization is utilized to recover sparse cell locations. We refer to our proposed cell detection framework as CNNCS (convolutional neural network + compressed sensing).

Output space encoding/representation/transformation sometimes yields more accurate predictions in machine learning \cite{ECOC}, \cite{RAkEL}. In the past, CS-based encoding was used in conjunction with linear and non-linear predictions \cite{CS}, \cite{Bayesian-CS}, \cite{output-space-thesis}. We believe, the proposed CNNCS is the first such attempt to solve cell detection and localization that achieved competitive results on benchmark datasets. There are several advantages of using CS-based output encoding for cell detection and localization. First, the compressed output vector is much shorter in length than the original sparse pixel space. So, the memory requirement would be typically smaller and consequently, there would be less risk of overfitting. Next, there are plenty of opportunities to apply ensemble averages to improve generalization. Furthermore, CS-theory dictates that pairwise distances are approximately maintained in the compressed space \cite{Alpher16}, \cite{Alpher17}. Thus, even after output space encoding, the machine learner still targets the original output space in an equivalent distance norm. From earlier research, we also point out a generalization error bound for such systems. Our contribution is summarized below:

First, this is the first attempt to combine deep learning with CS-based output encoding to solve cell detection and localization problem. 

Second, we try to overcome the aforementioned class imbalance issue by converting a classification problem into a regression problem, where sparse cell locations are distributed by a random projection into a fixed length vector as a target for the regression. 

Third, we introduce redundancies in the CS-based output encoding that are exploited by CNN to boost generalization accuracy in cell detection and localization. This redundancies also help to reduce false detections.

Fourth, we demonstrate that the proposed CNNCS framework achieves competitive results compared to the state-of-the-art methods on several benchmark datasets and challenging cell detection contests.

\section{Background and Related Work}
\label{bg}
\subsection{General Object Detection}

Prior to deep learning, general object detection pipeline consisted of feature extraction followed by classifiers or detectors. Detection had traditionally been addressed using the handcrafted features such as SIFT \cite{sift}, HOG \cite{HOG}, LBP \cite{LBP}, etc. At that time, progress in object detection greatly depended on the invention of more discriminative hand-crafted features. Following the big success of Convolutional Neural Network (CNN) in image classification task \cite{ImageNet}, deep learning model have been widely adopted in the computer vision community. For example, Region-based Convolutional Neural Networks (R-CNN) \cite{rcnn} and its variants \cite{fast-rcnn}, \cite{faster-rcnn}, Fully Convolutional Networks (FCN) \cite{FCN} with recent optimization \cite{redmon2016yolo9000} have become the state-of-the-art algorithms for the extended object detection problems.

%At the very begining, AlexNet \cite{ImageNet} architecture was widely adoptted and modified for various vision problems. After that, several milestone networks (VGGNet \cite{VGGnet}, ResNet \cite{ResNet}, Inception nets \cite{SzegedyIV16} and DenseNets \cite{huang2017densely} among many others) have been proposed. Deep learning has become one of the most influential methods in computer vision, and achieved successes in most major computer vision tasks like classification \cite{ImageNet}, \cite{ResNet}, \cite{huang2017densely}, detection \cite{rcnn}, \cite{fast-rcnn}, \cite{faster-rcnn}, \cite{redmon2016yolo9000} and semantic image \cite{FCN} and video segmentation \cite{Valipour2017}.

Once again, cell detection or localization task is not simply a sub-task of a general object detection, those state-of-the-art solutions for general object detection are able to provide useful clues to right direction, but cannot be easily translated to cell detection.

\subsection{Cell Detection and Localization}

The state-of-the-art methods in detection and localization today include deep learning techniques for cell detection and localization. Recently a deep Fully Convolutional Network (FCN) \cite{FCN} was proposed for the image segmentation problem and had shown remarkable performance. Soon after the FCN is proposed, \cite{Alpher25} presented a FCN-based framework for cell counting, where their FCN is responsible for predicting a spatial density map of target cells, and the number of cells can be estimated by an integration over the learned density map. Slightly similar to \cite{Alpher25}, a cascaded network \cite{CasNN} has been proposed for cell detection. \cite{CasNN} uses a FCN for candidate region selection, and then a CNN for further discrimination between target cells and background.

In \cite{DNN-IDSIA}, a mitosis detection method has been proposed by CNN-based prediction followed by ad-hoc post processing. As the winner of ICPR 2012 mitosis detection competition, \cite{DNN-IDSIA} used deep max-pooling convolutional neural networks to detect mitosis in breast histology images. The networks were trained to classify each pixel in the images, using a patch centered on the pixel as context. Then post processing was applied to the network output. In \cite{AggNet16}, expectation maximization has been utilized within deep learning framework in an end-to-end fashion for mitosis detection. This work presents a new concept for learning from crowds that handle data aggregation directly as part of the learning process of the convolutional neural network (CNN) via additional crowd-sourcing layer. It is the first piece of work where deep learning has been applied to generate a ground-truth labeling from non-expert crowd annotation in a biomedical context.

\subsection{Compressed Sensing}

During the past decade, compressed sensing or compressive sensing (CS) \cite{Alpher16} has emerged as a new framework for signal acquisition and reconstruction, and has received growing attention, mainly motivated by the rich theoretical and experimental results as shown in many reports \cite{Alpher17}, \cite{Alpher18}, \cite{Alpher16}, and so on. As we know, the Nyquist-Shannon sampling theorem states that a certain minimum sampling rate is required in order to reconstruct a band-limited signal. However, CS enables a potentially large reduction in the sampling and computation costs for sensing/reconstructing signals that are sparse or have a sparse representation under some linear transformation (e.g. Fourier transform).

Under the premise of CS, an unknown signal of interest is observed (sensed) through a limited number of linear observations. Many works \cite{Alpher17}, \cite{Alpher18}, \cite{Alpher16} have proven that it is possible to obtain a stable reconstruction of the unknown signal from these observations, under the general assumptions that the signal is sparse (or can be represented sparsely with respect to a linear basis) and matrix discoherence. The signal recovery techniques typically rely on a convex optimization with a penalty expressed by $\emph{L}_1$ norm, for example orthogonal matching pursuit (OMP) \cite{OMP} and dual augmented Lagrangian (DAL) method \cite{DAL}.

\section{Proposed Method}

\subsection{System Overview}

\begin{figure}[h]
	\centering
	\setlength{\abovecaptionskip}{0pt}
	\setlength{\belowcaptionskip}{0pt}
	\includegraphics[width=13cm]{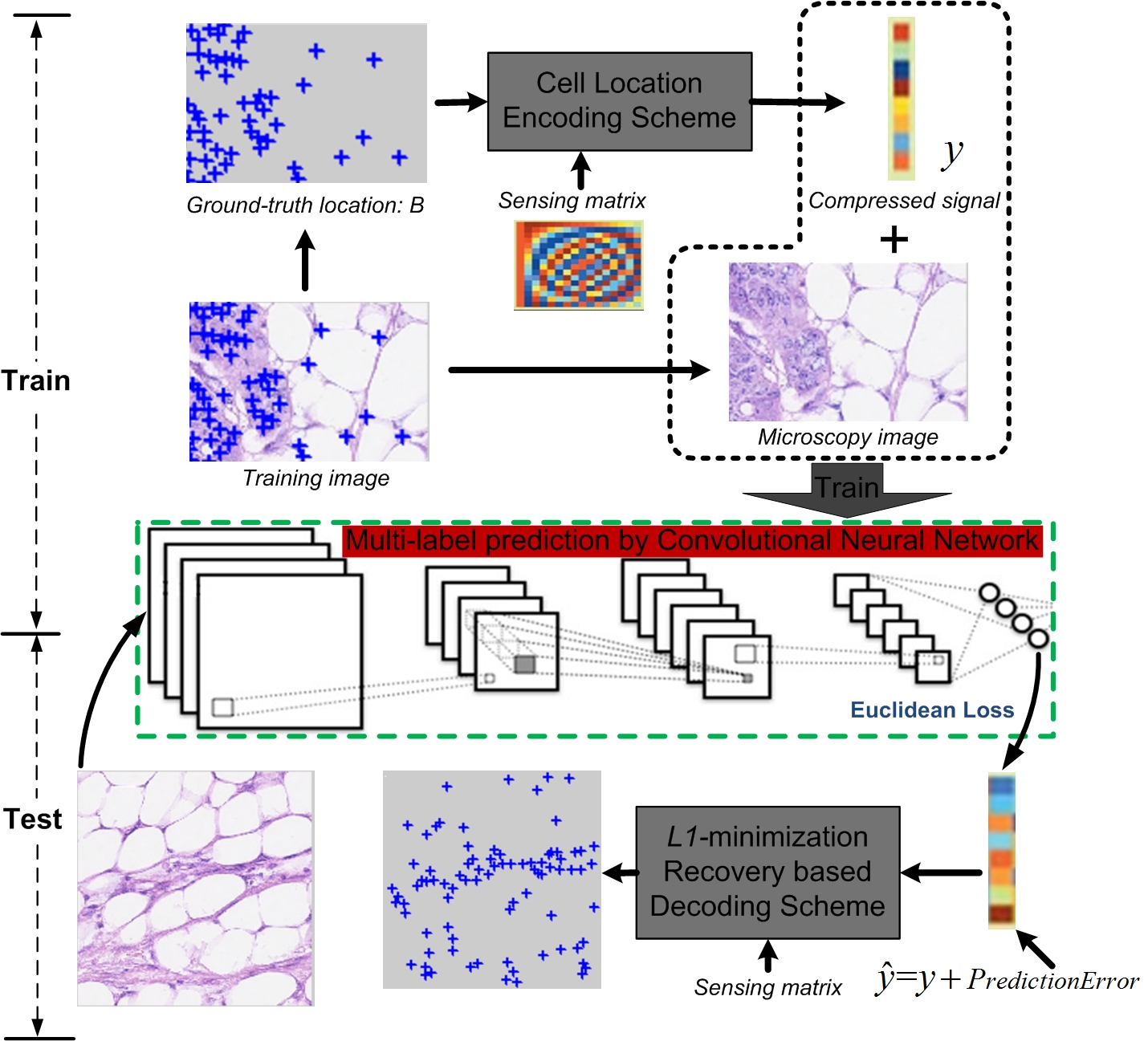}
	\caption{The system overview of the proposed CNNCS framework for cell detection and localization.}
	\label{baseline-system-overview}
\end{figure}

The proposed detection framework consists of three major components: (1) cell location encoding phase using random projection, (2) a CNN based regression model to capture the relationship between a cell microscopy image and the encoded signal $y$, and (3) decoding phase for recovery and detection. The flow chart of the whole framework is shown in Fig.~\ref{baseline-system-overview}.

During training, the ground truth location of cells are indicated by a pixel-wise binary annotation map $B$. We propose two cell location encoding schemes, which convert cell location from the pixel space representation $B$ to a compressed signal representation $y$. Then, training pairs, each consisting of a cell microscopy image and the compressed signal $y$, train a CNN to work as a multi-label regression model. We employ the Euclidean loss function during training, because it is often more suitable for a regression task. Image rotations may be performed on the training sets for the purpose of data augmentation as well as making the system more robust to rotations.

During testing, the trained network is responsible for outputting an estimated signal $\hat{y}$ for each test image. After that, a decoding scheme is designed to estimate the ground truth cell location by performing $L_1$ minimization recovery on the estimated signal $\hat{y}$, with the known sensing matrix.

\subsection{Cell Location Encoding and Decoding Scheme}

\subsubsection{Encoding Schemes}
In the CNNCS framework, we employ two types of random projection-based encodings as described below.

\textbf{Scheme-1: Encoding by Reshaping}

For the cell detection problem, cells are often annotated by pixel-level labels. The most common way is to attach a dot or cross at the center of every cell, instead of a bounding box around the cell. So, let us suppose there is a pixel-wise binary annotation map $B$ of size $w$-by-$h$, which indicates the location of cells by labeling 1 at the pixels of cell centroids, otherwise labeling 0 at background pixels. To vectorize the annotation map $B$, the most intuitive scheme is to concatenate every row of $B$ into a binary vector $f$ of length  $wh$. Thus, a positive element in $B$ with $\{x,y\}$ coordinates will be encoded to the $[x+h(y-1)]$-th position in $f$. $f$ is also a $k$-sparse signal, so, there are at most $k$ non-zero entries in $f$. Here, we refer this intuitive encoding scheme as "Scheme-1: Encoding by Reshaping".

After the vector $f$ is generated, we apply a random projection. CS theory guarantees that $f$ could be fully represented by linear observations $y$:
\begin{equation}
y=\Phi f,
\end{equation}
provided the sensing matrix $\Phi$ satisfies a restricted isometry property (RIP) condition \cite{Alpher17}, \cite{Alpher18}. In many cases, $\Phi$ is typically a $M \times N$ ($M \ll N=hw$) random Gaussian matrix. Here, the number of observations $M$ is much smaller than $N$, and obeys: $M \geq  C_M k log(N)$, where $C_M$ is a small constant greater than one.

\textbf{Scheme-2: Encoding by Signed Distances}

For the encoding scheme-1, the space complexity of the interim result $f$ is $\mathcal{O}(wh)$. For example, to encode the location of cells in a 260-by-260 pixel image, scheme-1 will produce $f$ as a 67,600-length vector; so that in the subsequent CS process, a huge sensing matrix in size of $M$-by-67600 is required in order to match the dimension of $f$, which will make the system quite slow, even unacceptable for larger images. To further optimize the encoding scheme, we propose a second scheme, where the coordinates of every cell centroid are projected onto multiple observation axes. We refer the second encoding scheme as "Scheme-2: Encoding by Projection."

\begin{figure*}[htbp]
	\centering
	\setlength{\abovecaptionskip}{0pt}
	\setlength{\belowcaptionskip}{0pt}
	\includegraphics[width=17cm]{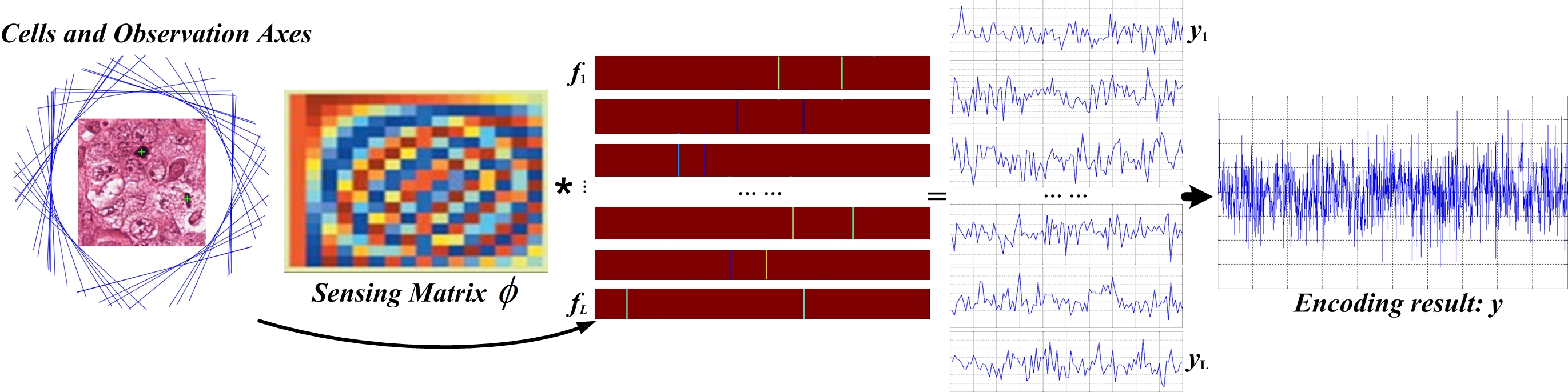}
	\caption{Cell location encoding by signed distances (Scheme-2).}
	\label{encoding}
\end{figure*}

To encode location of cells, we create a set of observation axes $OA=\left\lbrace oa_{l} \right\rbrace, l=1,2,\dots, L$, where $L$ indicates the total number of observation axes used. The observation axes are uniformly-distributed around an image (See Fig.~\ref{encoding}, left-most picture) For the $l$-th observation axis $oa_{l}$, the location of cells is encoded into a $R$-length ($R=\sqrt{w^{2}+h^{2}}$) sparse signal, referred as $f_{l}$ (See Fig.~\ref{encoding}, third picture). We calculate the perpendicular signed distances ($f_{l}$) from cells to $oa_{l}$. Thus, $f_{l}$ contains signed distances, which not only measure the distance, but also describe on which side of $oa_{l}$ cells are located. After that, the encoding of cell locations under $oa_{l}$ is $y_{l}$, which is obtained by the following random projection:
\begin{equation}
y_{l}=\Phi f_{l},
\end{equation}
We repeat the above process for all the $L$ observation axes and obtain each $y_{l}$. After concatenating all the $y_{l}$, $l=1,2,\dots, L$, the final encoding result $y$ is available, which is the joint representation of cells location. The whole encoding process is illustrated by Fig.~\ref{encoding}.

For encoding scheme-2, the size of the sensing matrix $\Phi$ is $M$-by-$\sqrt{w^{2}+h^{2}}$. In comparison, encoding scheme-1 requires a much larger sensing matrix of size $M$-by-$wh$. The first advantage of encoding scheme-2 is that it dramatically reduces the size of the sensing matrix, which is quite helpful for the recovery process, especially when the size of images is large. Secondly, the encoding result $y$ carries \textbf{redundant} information about cell locations. In the subsequent decoding phase, averaging over the redundant information makes the final detection more reliable. More details can be found in experiments section. A final point is that in case more than one cell locations are projected to the same bin in a particular observation axis, such a conflict will not occur for the same set of cells at other observation axes.

\subsubsection{Decoding Scheme}

Accurate recovery of $f$ can be obtained from the encoded signal $y$ by solving the following $L_1$ norm convex optimization problem:
\begin{equation}
\hat{f}=\arg\min_f \| f \|_1 \qquad \text{subject to}\qquad y=\Phi f
\end{equation}
After $\hat{f}$ is recovered, every true cell is localized $L$ times, i.e. with $L$ candidate positions predicted. The redundancy information allows us to estimate more accurate detection of a true cell.

\begin{figure}[h]
	\centering
	\setlength{\abovecaptionskip}{0pt}
	\setlength{\belowcaptionskip}{0pt}
	\includegraphics[width=13cm]{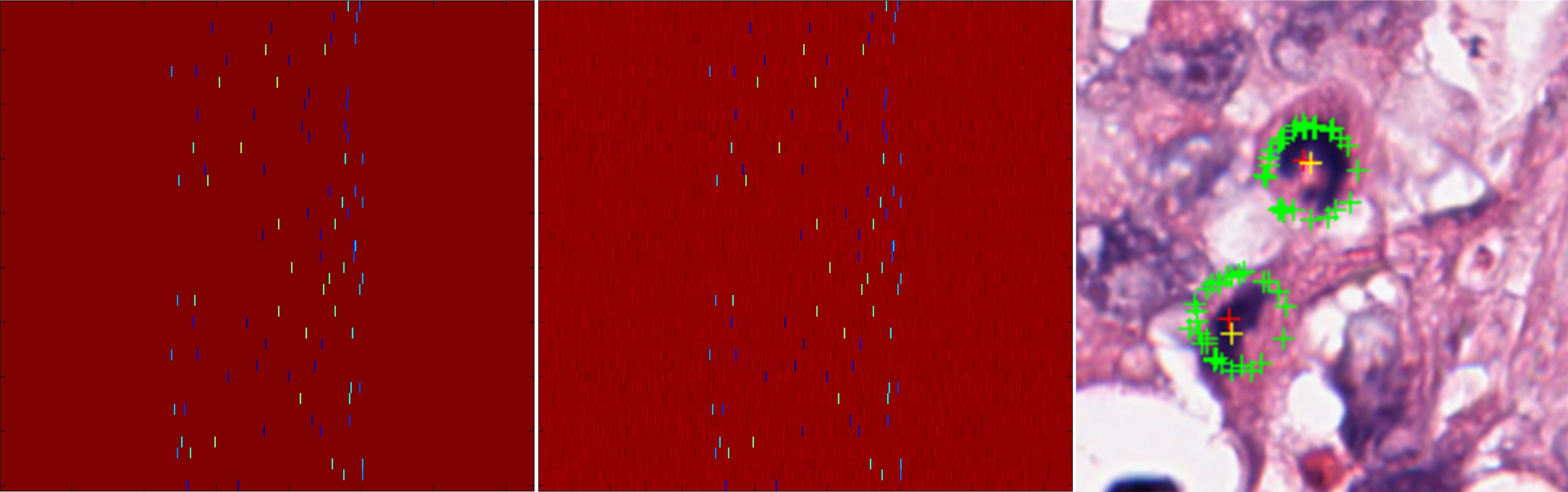}
	\caption{Cell Location Decoding Scheme. From left to right: true location signal $f$, decoded location signal $\hat{f}$ and detection results. Yellow crosses indicate the ground-truth location of cells, green crosses are the candidates points, red crosses represent the final detected points.}
	\label{decoding}
\end{figure}

The first two images of Fig.~\ref{decoding} from left present examples of the true location signal $f$ and decoded location signal $\hat{f}$, respectively. The noisy signed distances of $\hat{f}$ are typically very close to each observation axis. That is why we create observation axes outside of the image space, so that these noisy distances can be easily distinguished from true candidate distances. This separation is done by mean shift clustering, which also groups true detections into localized groups of detections. Two such groups (clusters) are shown in Fig.~\ref{decoding}, where the signed distances formed circular patterns of points (in green) around ground truth detections (in yellow). Averaging over these green points belonging to a cluster provides us a predicted location (in red) as shown in Fig.~\ref{decoding}.

\subsection{Signal Prediction by Convolutional Neural Network}

\begin{figure*}[htbp]
	\centering
	\setlength{\abovecaptionskip}{0pt}
	\setlength{\belowcaptionskip}{0pt}
	\includegraphics[width=17cm]{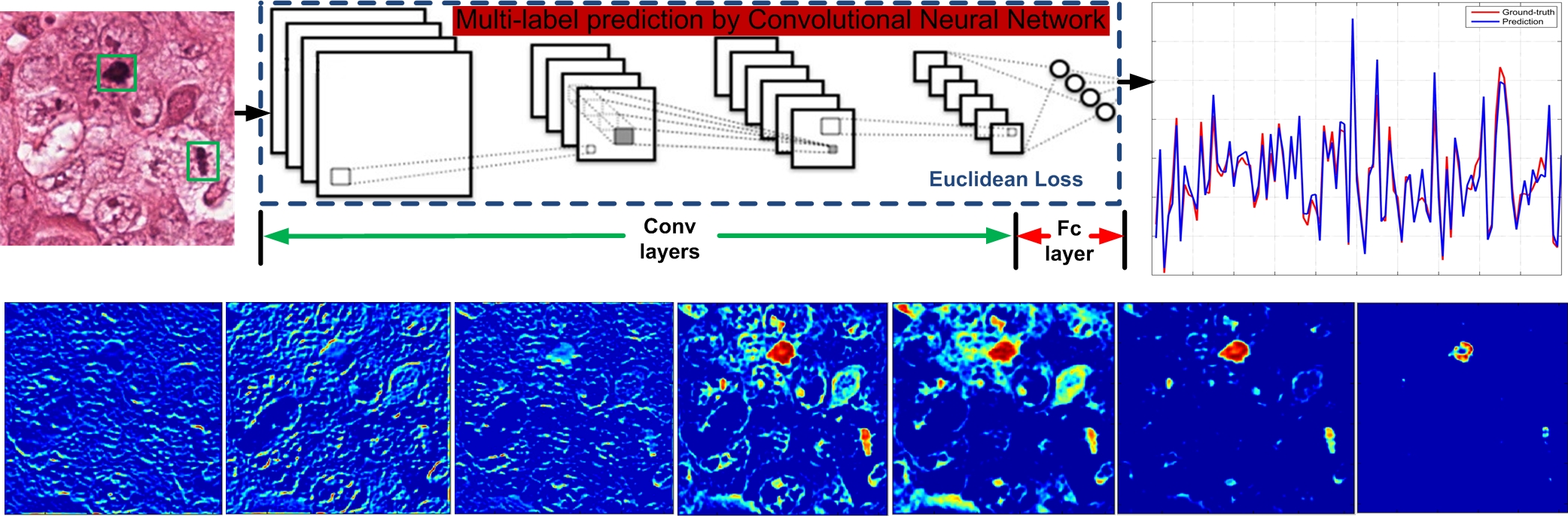}
	\caption{An illustration of the process of signal prediction by convolutional neural network. The bottom row presents the feature maps learned from Convolutional (Conv) layers of the CNN with training process going on. The current CNN follows the AlexNet architecture. These feature maps come from the Conv1, Conv1, Conv2, Conv3, Conv3, Conv4 and Conv5 respectively. The top-right picture shows the ground-truth compressed signal (red) and compressed signal (blue) predicted from the Fullly-connected (Fc) layer of the CNN. From the picture, we can observe that the predicted signal approximate the pattern of ground truth signal well.}
	\label{heatmap}
\end{figure*}

We utilize a CNN to build a regression model between a cell microscopy image and its cell location representation: compressed signal $y$. We employ two kinds of CNN architectures. One of them is AlexNet \cite{ImageNet}, which consists of 5 convolution layers + 3 fully connected layers; the other is the deep residual network (ResNet) \cite{ResNet} where we use its 152-layer model. In both the architectures, the loss function is defined as the Euclidean loss. The dimension of output layer of AlexNet and ResNet has been modified to the length of compressed signal $y$. We train the AlexNet model from scratch, in comparison, we perform fine-tuning on the weights in fully-connected layer of the ResNet.

To prepare the training data, we generate a large number of square patches from training images. Along with each training patch, there is a signal (i.e. the encoding result: $y$), which indicates the location of target cells present in the patch. After that, patch rotation is performed on the collected training patches for data augmentation and making the system rotation invariant.

The trained CNN not only predicts the signal from its output layer, the feature maps learned from its Conv layers also provide rich information for recognition. Fig.~\ref{heatmap} visualizes the learned feature maps, which represents the probabilistic score or activation maps of target cell regions (indicated by green boxes in the left image) during training process. It can be observed that higher scores are fired on the target regions of score masks, while most of the non-target regions have been suppressed more and more with training process going on.

To further optimize our CNN model, we apply Multi-Task Learning (MTL) \cite{Caruana1997}. During training a CNN, two kinds of labels are provided. The first kind is the encoded vector: $y$, which carries the pixel-level location information of cells. The other kind is a scalar: cell count ($c$), which indicates the total number of cells in a training image patch. We concatenate the two kinds of labels into the final training label by $label = \left\lbrace y,\lambda c\right\rbrace$, where $\lambda$ is a hyper parameter. Then, Euclidean loss is applied on the fusion label. Thus, supervision information for both cell detection and cell counting can be jointly used to optimize the parameters of our CNN model.

\subsection{Theoretical Justification}

\textbf{Equivalent Targets for Optimization}

We first show that from the optimization standpoint, compressed vector is a good proxy for the original, sparse output space. This result directly follows from the CS theory. As mentioned before, $f$ indicates the cell location represented in pixel space, and $y$ is the cell location represented in compressed signal space. They follow the relationship: $y=\Phi f$, where $\Phi$ is the sensing matrix. Let us assume that $f_p$ and $f_g$ are respectively the prediction and ground-truth vectors in the pixel space. Similarly, we have $y_p$ and $y_g$ as their compressed counterparts, respectively. 

\textbf{Claim:} $\left \| y_{g}-y_{p} \right \|$ and $\left \| f_{g}-f_{p} \right \|$ are approximately equivalent targets for optimization. 

\textbf{Proof:} According to the CS theory, a sensing matrix $\Phi \in \mathbb{R}^{m\times d}$ should satisfy the $\left( k,\delta \right)-restricted\ isometry\ property \left(\left(k,\delta \right)-RIP\right)$, which states that for all $k-$sparse $f\in\mathbb{R}^{d}$, $\delta\in\left( 0,1\right)$, the following holds \cite{Alpher17}, \cite{Alpher18}, \cite{Alpher16}:
\begin{equation}
\left(1-\delta \right)\|f\|\leq\|\Phi f\|\leq\left(1
+\delta \right)\|f\|.
\end{equation}

Note that if the sensing matrix $\Phi$ satisfies $(2k,\delta)$-RIP, then (4) also holds good. Now replace $f$ with $\left(f_{g}-f_{p}\right)$ and note that $\left(f_{g}-f_{p}\right) $ is $2k$-sparse. Thus,
\begin{equation}
\left(1-\delta \right)\left \| f_{g}-f_{p} \right \|\leq \left \|y_{g}-y_{p} \right \|\leq \left(1+\delta \right)\left \| f_{g}-f_{p} \right \|.
\end{equation}
From the right hand side inequality, we note that if $\left \| f_{g}-f_{p} \right \|$ is small, then $\left \| y_{g}-y_{p} \right \|$ would be small too. In the same way, if $\left \| y_{g}-y_{p} \right \|$ is large, then the inequality implies that $\left \| f_{g}-f_{p} \right \|$ would be large too. Similarly, from the left hand side inequality, we note that if $\left \| f_{g}-f_{p} \right \|$ is large then $\left \| y_{g}-y_{p} \right \|$ will be large, and if $\left \| y_{g}-y_{p} \right \|$ is small then $\left \| f_{g}-f_{p} \right \|$ will small too. These relationships prove the claim that from the optimization perspective $\left \| y_{g}-y_{p} \right \|$ and $\left \| f_{g}-f_{p} \right \|$ are approximately equivalent.\\

\noindent\textbf{A Bound on Generalization Prediction Error}\\
In this section we mention a powerful result from \cite{CS}. Let $h$ be the predicted compressed vector by the CNN, $f$ be the ground truth sparse vector, $\hat{f}$ be the reconstructed sparse vector from prediction, and $\Phi$ be the sensing matrix. Then the generalization error bound provided in \cite{CS} is as follows:
\begin{equation}
\|\hat{f}-f \|^{2}_{2}\leq C_{1}\cdot\|h-\Phi f \|^{2}_{2}+C_{2}\cdot sperr(\hat{f},f),
\end{equation}
where $C_1$ and $C_2$ are two small constants and $sperr$ measures how well the reconstruction algorithm has worked \cite{CS}. This result demonstrates that expected error in the original space is bound by the expected errors of the predictor and that of the reconstruction algorithm. Thus, it makes sense to apply a very good machine learner such as deep CNN that can minimize the first term in the right hand side of (6). On the other hand, DAL provides one of the best $L_1$ recovery algorithms to minimize the second term in the right side of (6).

\section{Experiments}

\subsection{Datasets and Evaluation Criteria}
%\label{ssec:subhead}

We utilize seven cell datasets, on which CNNCS and other comparison methods are evaluated. The 1st dataset \cite{Nuclei-data} involves 100 H$\&$E stained histology images of colorectal adenocarcinomas. The 2nd dataset \cite{bacterial-data} consists of 200 highly realistic synthetic emulations of fluorescence microscopic images of bacterial cells. The 3rd dataset \cite{our-ECCV} comprises of 55 high resolution microscopic images of breast cancers double stained in red (cytokeratin – epithelial marker) and brown (nuclear – proliferative marker). The 4th dataset is the ICPR 2012 mitosis detection contest dataset \cite{ICPR-2012} including 50 high-resolution (2084-by-2084) RGB microscope slides of Mitosis. The 5th dataset \cite{ICPR-2014} is the ICPR 2014 grand contest of mitosis detection, which is a follow-up and an extension of the ICPR 2012 contest on detection of mitosis. Compared with the contest in 2012, the ICPR 2014 contest is much more challenging, which contains way more images for training and testing. The 6th dataset is the AMIDA-2013 mitosis detection dataset \cite{AMIDA-2013}, which contains 676 breast cancer histology images belonging to 23 patients. The 7th dataset is the AMIDA-2016 mitosis detection dataset \cite{AMIDA-2016}, which is an extension of the AMIDA 2013 contest on detection of mitosis. It contains 587 breast cancer histology images belonging to 73 patients for training, and 34 breast cancer histology images for testing with no ground truth available. For each dataset, the annotation that represents the location of cell centroids is shown in Fig.\ref{dataset}, details of datasets are summarized in Table.~\ref{data-table}.

\begin{table}[h] \footnotesize
	\setlength{\abovecaptionskip}{0pt}
	\setlength{\belowcaptionskip}{0pt}
	\begin{center}
		\caption{\emph{Size} is the image size; \emph{Ntr/Nte} is the number of images selected for training and testing; \emph{AC} indicates the average number of cells per image.}
		\label{data-table}
		\begin{tabular}{*{22}{c}}
			\hline\noalign{\smallskip}
			Cell Dataset & Size & Ntr/Nte & AC\\
			\noalign{\smallskip}
			\hline
			\noalign{\smallskip}
			
			Nuclei \cite{Nuclei-data}& 500$\times$500 &  50/50 & 310.22\\
			
			Bacterial \cite{bacterial-data} & 256$\times$256 & 100/100 & 171.47\\
			
			Ki67 Cell \cite{our-ECCV} & 1920$\times$2560 & 45/10 & 2045.85\\
			
			ICPR 2012 \cite{ICPR-2012} & 2084$\times$2084 & 35/15 & 5.31\\
			
			ICPR 2014 \cite{ICPR-2014} & 1539$\times$1376 & 1136/496 & 4.41\\
			
			AMIDA 2013 \cite{AMIDA-2013} & 2000$\times$2000 & 447/229 & 3.54\\
			
			AMIDA 2016 \cite{AMIDA-2016} & 2000$\times$2000 & 587/34 & 2.13\\
			
			\hline
		\end{tabular}
	\end{center}
\end{table}

\begin{figure}[h]
	\centering
	\setlength{\abovecaptionskip}{0pt}
	\setlength{\belowcaptionskip}{0pt}
	\includegraphics[width=13cm]{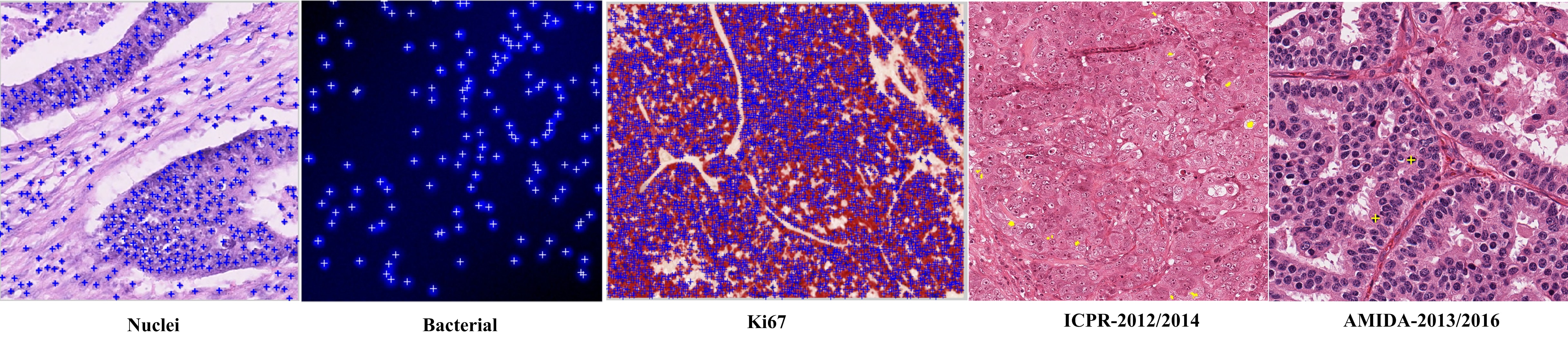}
	\caption{Dataset examples and their annotation.}
	\label{dataset}
\end{figure}

For evaluation, we adopt the criteria of the ICPR 2012 mitosis detection contest \cite{ICPR-2012}, which is also adopted in several other cell detection contests. A detection would be counted as true positive ($TP$) if the distance between the predicted centroid and ground truth cell centroid is less than $\rho$. Otherwise, a detection is considered as false positives ($FP$). The missed ground truth cells are counted as false negatives ($FN$). In our experiments, $\rho$ is set to be the radius of the smallest cell in the dataset. Thus, only centroids that are detected to lie inside cells are considered correct. The results are reported in terms of Precision: $P=TP/(TP+FP)$ and Recall: $R=TP/(TP+FN)$ and $F_1$-score: $F_1=2PR/(P+R)$ in the following sections.

\subsection{Experiments with Encoding Scheme-1}

To evaluate, we carry out performance comparison experiment between CNNCS and three state-of-the-art cell detection methods (``FCN-based'' \cite{Alpher25}, ``Le.detect'' \cite{Alpher21}, ``CasNN'' \cite{CasNN}). In this experiment, the scheme-1: encoding by reshaping is applied in CNNCS.

\begin{figure}[h]
	\centering
	\setlength{\abovecaptionskip}{0pt}
	\setlength{\belowcaptionskip}{0pt}
	\includegraphics[width=8cm]{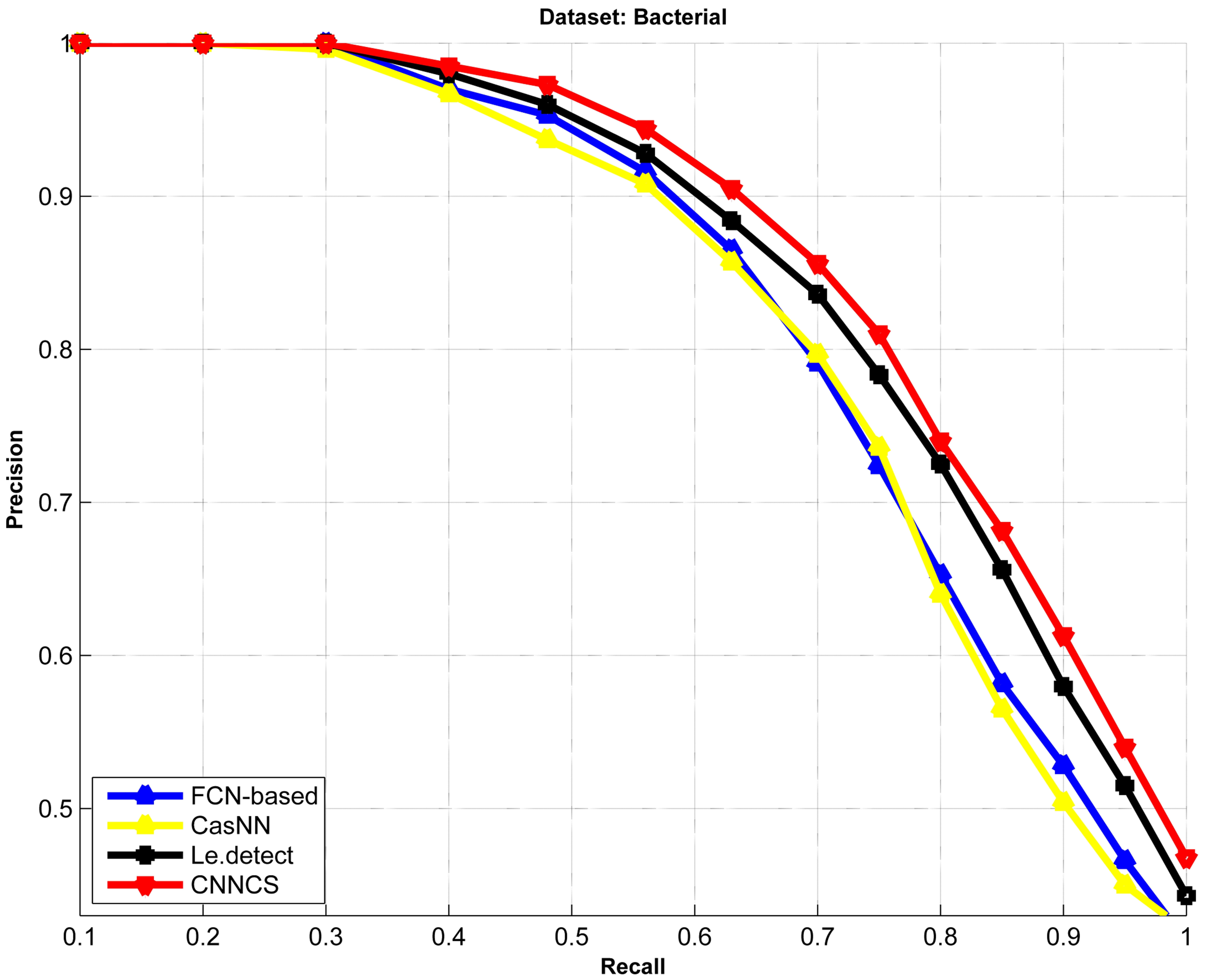}
	\includegraphics[width=8cm]{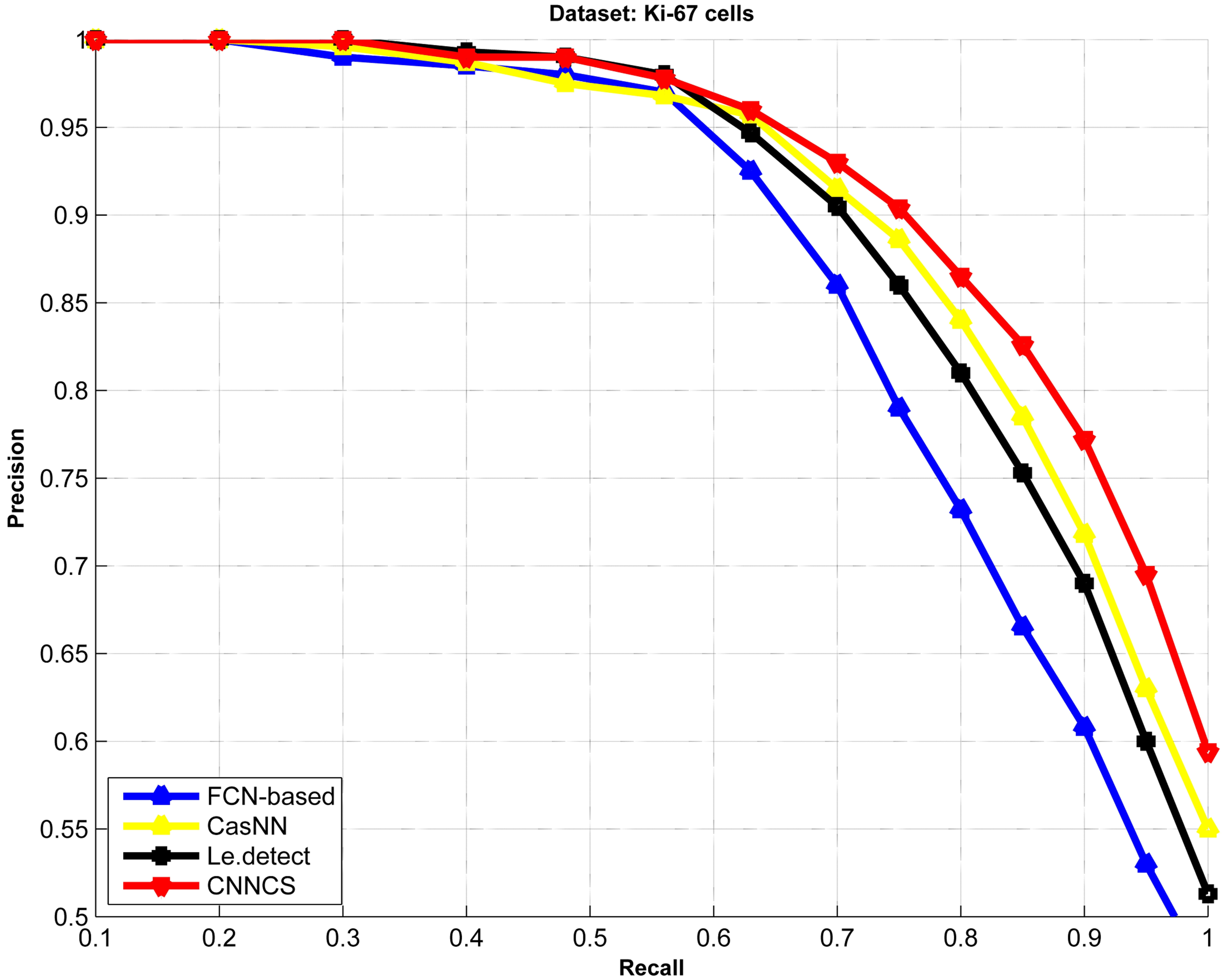}
	\includegraphics[width=8cm]{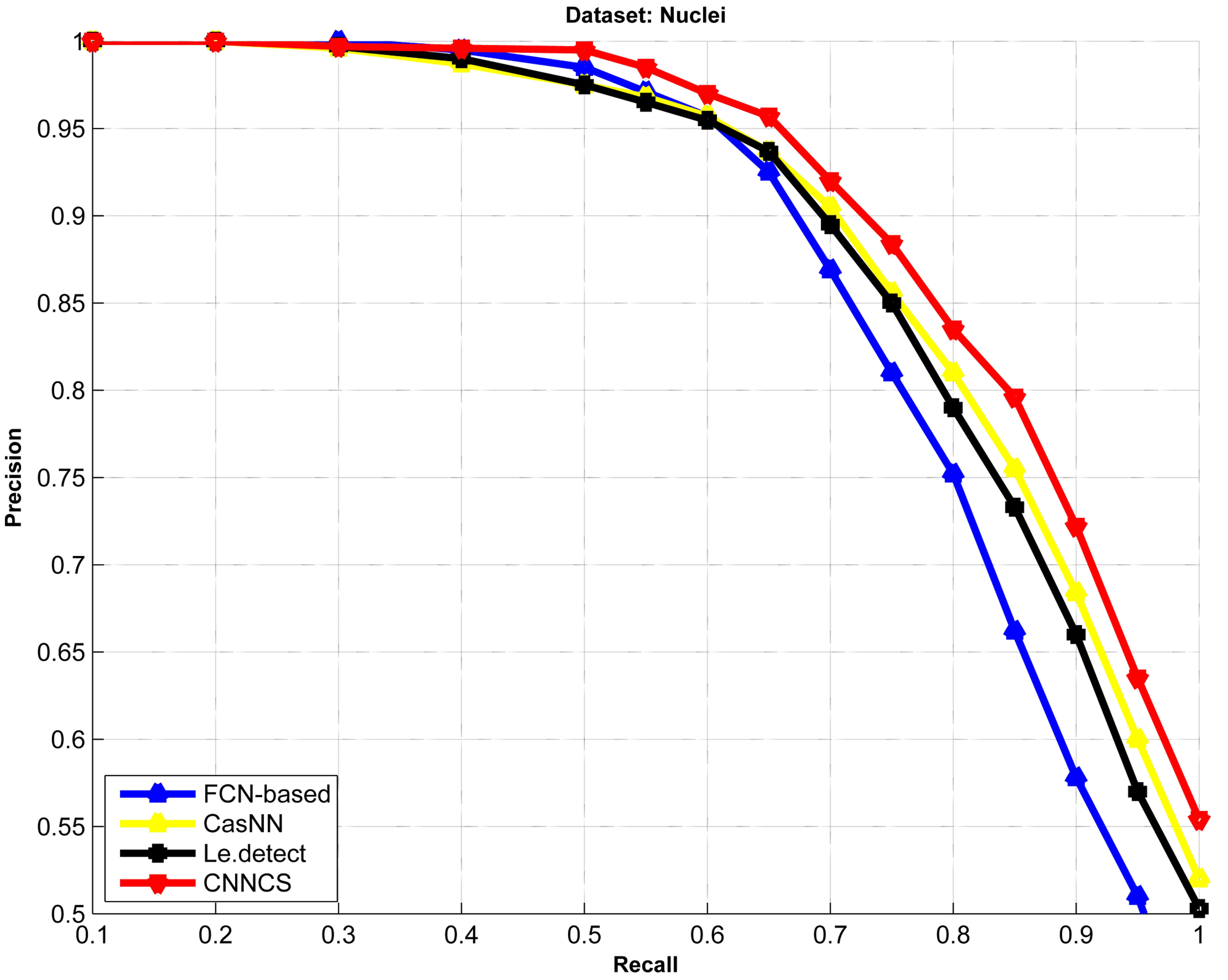}
	\caption{Precision and recall curves of four methods on three datasets.}
	\label{per}
\end{figure}

For the four methods to provide different values of Precision-Recall as shown in Fig.~\ref{per}, we tune hyper parameters of every method. With scheme-1, CNNCS has a threshold $T$ to apply on the recovered sparse signal $\hat{f}$ before re-shaping it to a binary image $B$. $T$ is used to perform cell vs. non-cell binary classification and can be treated as a hyper parameter during training. In ``FCN-based'' \cite{Alpher25}, there is also a threshold applied to the local probability-maximum candidate points to make final decision about cell or non-cell. Similarly, in the first step of ``Le.detect'' \cite{Alpher21}, researchers use a MSER-detector (a stability threshold involved here) to produce a number of candidate regions, on which their learning procedure determines which of these candidates regions correspond to cells. In the first experiment, we analyze the three methods using Precision-Recall curves by varying their own thresholds.

Fig.~\ref{per} presents Precision-Recall curves on three cell datasets. All the four methods give reliable detection performances in the range of recall=[0.1-0.4]. After about recall=0.6, the precision of ``FCN-based'' \cite{Alpher25} drops much faster. This can be attributed to the fact that ``FCN-based'' \cite{Alpher25} works by finding local maximum points on a cell density map. However, the local maximum operation fails in several scenarios, for example when two cell density peaks are close to each other, or large peak may covers neighboring small peaks. Consequently, to obtain the same level of recall, ``FCN-based'' \cite{Alpher25} provides many false detections.

Furthermore, it also can be observed that CNNCS has an improvement over ``Le.detect'' \cite{Alpher21} (red line clearly outperforms black line under varying recall values). This can be largely explained by the fact that traditional methods (no matter if \cite{Alpher21} or \cite{Alpher25} is used) always try to predict the coordinates of cells directly on a 2-D image. The coordinates are sensitive to system prediction bias or error, considering the nature of cell detection that cells are small and quite dense in most cases. It is not surprising that ``Le.detect'' \cite{Alpher21} will miss some cells and/or detect other cells in wrong locations. In comparison, CNNCS transfers the cell detection task from pixel space to compressed signal space, where the location information of cells is no longer represented by $\left\lbrace x,y\right\rbrace $-coordinates. Instead, CNNCS performs cell detection by regression and recovery on a fixed length compressed signal. Compared to $\left\lbrace x,y\right\rbrace$-coordinates representation, the compressed signal is more robust to system prediction errors. For example, as shown in the right top corner of Fig.~\ref{heatmap}, even though there are differences between the ground-truth compressed signal and predicted compressed signal, the whole system can still give reliable detection performance as shown in Fig.~\ref{per}.

\begin{figure}[h]
	\centering
	\setlength{\abovecaptionskip}{0pt}
	\setlength{\belowcaptionskip}{0pt}
	\includegraphics[width=13cm]{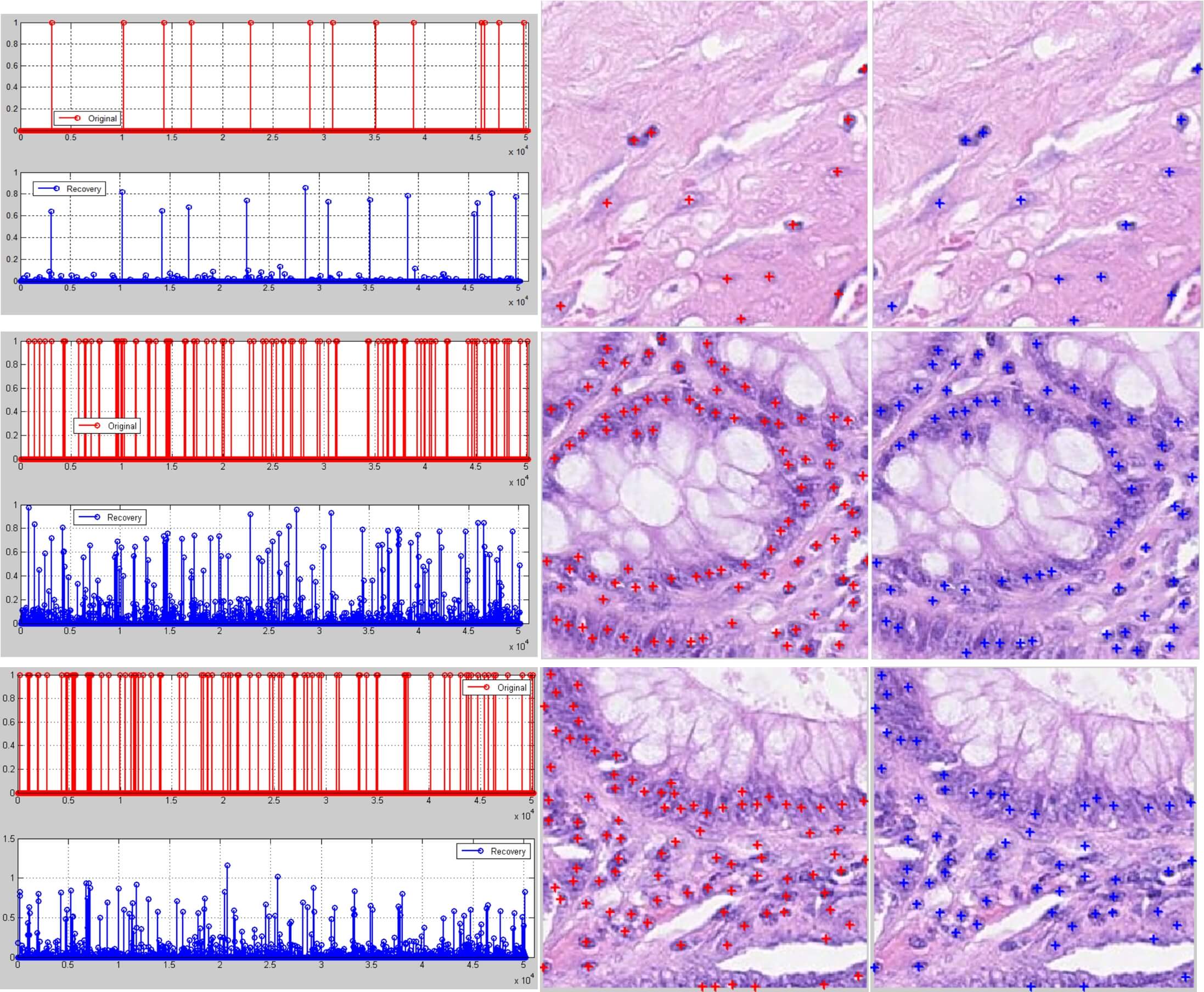}
	\caption{Detection results. Ground-truth: red, Prediction: blue. Left part shows the ground truth signal and the predicted sparse signal that carries the location information of cells; right part shows the ground-truth and detected cells.}
	\label{baseline-example}
\end{figure}

To get a better idea of the CNNCS method, we visualize a set of cell images with their detected cells and ground-truth cells in Fig.~\ref{baseline-example}. It can be observed that CNNCS is able to accurately detect most cells under a variety of conditions.

\subsection{Experiments with Encoding Scheme-2}

\subsubsection{Experiment on ICPR 2012 mitosis detection dataset}

To evaluate the performance of encoding scheme 2, we carry out the second group of performance comparison experiments. In the first experiment, we apply the proposed method on the ICPR 2012 mitosis detection contest dataset, which consists of 35 training images and 15 testing images. For the training process, we extracted image sub-samples (260-by-260) with no overlap between each other from the 35 training images. After that every 90$^\circ$ image rotation is performed on each sub-sample for data augmentation, resulting in a total of 8,960 training dataset. In addition, we perform random search to tune the three hyper parameters in scheme-2: (1) the number of rows in sensing matrix: $M$, (2) the number of observation lines: $L$ and (3) the importance ($\lambda$) of cell count during MTL. After that, the best performance is achieved when $M=112, L=27, \lambda=0.20$. Furthermore, we trained five CNN models to reduce the performance variance introduced by a single model and to improve the robustness of the whole system. Recently, deep residual network (ResNet) introduces residual connections into deep convolutional networks and has yielded state-of-the-art performance in the 2015 ILSVRC challenge \cite{ResNet}. This raises the question of whether there is any benefit in introducing and exploiting more recent CNN architectures into the cell detection task. Thus, in the experiment, we have explored the performance of CNNCS with different neural network architectures (AlexNet and ResNet). Finally, CNNCS gets the \textbf{highest} F1-score among all the comparison methods, details are summarized in Table~\ref{ICPR-2012-table}.

\begin{table}[h] \footnotesize
	\setlength{\abovecaptionskip}{0pt}
	\setlength{\belowcaptionskip}{0pt}
	\begin{center}
		\caption{Results of ICPR 2012 grand challenge of mitosis detection.}
		\label{ICPR-2012-table}
		\begin{tabular}{*{22}{c}}
			\hline\noalign{\smallskip}
			Method & Precision & Recall & F$_1$-score \\
			\noalign{\smallskip}
			\hline
			\noalign{\smallskip}
			
			UTRECHT & 0.511 & 0.680 & 0.584\\
			
			NEC \cite{NEC} & 0.747 & 0.590 & 0.659\\
			
			IPAL \cite{IPAL} & 0.698 & 0.740 & 0.718\\
			
			DNN \cite{DNN-IDSIA} &0.886&0.700&0.782\\
			
			RCasNN \cite{CasNN} &0.720&0.713&0.716\\
			
			CasNN-single \cite{CasNN} &0.738&0.753&0.745\\
			
			CasNN-average \cite{CasNN} &0.804&0.772&0.788\\
			
			\noalign{\smallskip}
			\cdashline{1-4}[11pt/3pt]
			\noalign{\smallskip}
			
			CNNCS-AlexNet &0.860&0.788&0.823\\
			
			CNNCS-ResNet &0.867&0.801&0.833\\
			
			CNNCS-ResNet-MTL &0.872&0.805&{\bf0.837}\\
			
			\hline
		\end{tabular}
	\end{center}
\end{table}

Compared to the state-of-the-art method: CasNN-average \cite{CasNN}, CNNCS with ResNet and MTL achieved a better performance with $F_1$-score 0.837. It can be observed from Table~\ref{ICPR-2012-table} that the precision of our method outperforms the previous best precision by 0.06-0.07, and recall also has recorded about 0.02 improvement. This phenomenon can be attributed to the detection principle of our method, where every ground-truth cell is localized with multiple candidate points guaranteed to be around the true location, then the average coordinates of these candidates is computed as the final detection. As a result, localization closer to the true cell becomes more reliable compared to other methods, thus leading to a higher precision. In addition, an improvement of $F_1$-score from 0.833 to 0.837 achieved by MTL demonstrates that the knowledge jointly learned from cell detection and cell counting provides further benefits at negligible additional computations.

\subsubsection{Experiment on ICPR 2014 mitosis detection dataset}

In the second experiment, we evaluated CNNCS on the ICPR 2014 contest of mitosis detection dataset (also called MITOS-ATYPIA-14), which is a follow-up and an extension of the ICPR 2012 contest on detection of mitosis. Compared with the contest in 2012, the ICPR 2014 contest is much more challenging, which contains more images for training and testing. It provides 1632 breast cancer histology images, 1136 images for training, 496 images for testing. Each image is in the size of 1539$\times$1376. We divide the training images into training set (910 images) and validation set (226 images). We perform random search on the validation set to optimize the hyper parameters. The best performance on MITOS-ATYPIA-14 dataset is achieved when $M=103, L=30, \lambda=0.24$. On the test dataset, we have achieved the \textbf{highest} F1-score among all the participated teams. The F1-score of all the participated teams are shown in Table~\ref{ICPR-2014-table}. As we see, the CNNCS method shows significant improvement compared to the results of other teams in all the histology slice groups. On an average, CNNCS has almost doubled the F1-score of teams at the second place.

\begin{table}[h] \footnotesize
	\setlength{\abovecaptionskip}{0pt}
	\setlength{\belowcaptionskip}{0pt}
	\begin{center}
		\caption{Results of ICPR 2014 contest of mitosis detection in breast cancer histological images. F1-scores of participated teams are shown.}
		\label{ICPR-2014-table}
		\begin{tabular}{*{22}{c}}
			\hline\noalign{\smallskip}
			Slice group & CUHK & MINES & YILDIZ & STRAS & CNNCS \\
			\noalign{\smallskip}
			\hline
			\noalign{\smallskip}
			
			A06 &0.119&0.317&0.370&0.160&{\bf0.783}\\
			
			A08 &0.333&0.171&0.172&0.024&{\bf0.463}\\
			
			A09 &0.593&0.473&0.280&0.072&{\bf0.660}\\
			
			%A13 &0.000&0.000&0.015&0.000&{\bf0.000}\\
			
			A19 &0.368&0.137&0.107&0.011&{\bf0.615}\\
			
			Average &0.356&0.235&0.167&0.024&{\bf0.633}\\
			
			\hline
		\end{tabular}
	\end{center}
\end{table}

\subsubsection{Experiment on AMIDA 2013 mitosis detection dataset}

The third experiment was performed on the AMIDA-2013 mitosis detection dataset, which contains 676 breast cancer histology images, belonging to 23 patients. Suspicious breast tissue is annotated by at least two expert pathologists, to label the center of each cancer cell. We train the proposed CNNCS method using 377 images, validate on 70 training images and test it on the testing set of AMIDA-2013 challenge that has 229 images from the last 8 patients. We employ ResNet as the network architecture with data balancing and MTL in the training set. Similar to previous experiments, we perform random search on the validation set to optimize the hyper parameters. The best performance on AMIDA-2013 dataset is achieved when $M=118, L=25, \lambda=0.32$. Finally among all the 17 participated teams, we achieve the \textbf{third highest} F1-score=0.471, which is quite close to the second place, and has a significant improvement over the fourth place method \cite{AggNet16}. For details, Table~\ref{AMIDA-2013-table} summarizes the comparison between CNNCS and other methods.

\begin{table}[h] \footnotesize
	\setlength{\abovecaptionskip}{0pt}
	\setlength{\belowcaptionskip}{0pt}
	\begin{center}
		\caption{Results of AMIDA-2013 MICCAI grand challenge of mitosis detection.}
		\label{AMIDA-2013-table}
		\begin{tabular}{*{22}{c}}
			\hline\noalign{\smallskip}
			Method & Precision & Recall & F$_1$-score \\
			\noalign{\smallskip}
			\hline
			\noalign{\smallskip}
			
			IDSIA \cite{DNN-IDSIA} &0.610&0.612&0.611\\
			
			DTU &0.427&0.555&0.483\\
			
			AggNet \cite{AggNet16} &0.441&0.424&0.433\\
			
			CUHK &0.690&0.310&0.427\\
			
			SURREY &0.357&0.332&0.344\\
			
			ISIK &0.306&0.351&0.327\\
			
			PANASONIC &0.336&0.310&0.322\\
			
			CCIPD/MINDLAB	&0.353&0.291&0.319\\
			
			WARWICK	&0.171&0.552&0.261\\
			
			POLYTECH/UCLAN	&0.186&0.263&0.218\\
			
			MINES	&0.139&0.490&0.217\\
			
			SHEFFIELD/SURREY &0.119&0.107&0.113\\
			
			SEOUL	&0.032&0.630&0.061\\
			
			UNI-JENA	&0.007&0.077&0.013\\
			
			NIH	&0.002&0.049&0.003\\
			
			\noalign{\smallskip}
			\cdashline{1-4}[11pt/3pt]
			\noalign{\smallskip}
			
			CNNCS &0.3588&0.5529&0.4352\\
			
			\hline
		\end{tabular}
	\end{center}
\end{table}

\subsubsection{Experiment on AMIDA 2016 mitosis detection dataset}

In the fourth experiment, we participated in the AMIDA-2016 mitosis detection challenge (also called TUPAC16), which is a follow-up and an extension of the AMIDA-2013 contest on detection of mitosis. Its training dataset has 587 breast cancer histology images in size of 2000$\times$2000, belonging to 73 patients. Its test dataset contains 34 breast cancer histology images in the same size without publicly available ground truth labels.

\begin{table}[h] \footnotesize
	\setlength{\abovecaptionskip}{0pt}
	\setlength{\belowcaptionskip}{0pt}
	\begin{center}
		\caption{Results of AMIDA-2016 MICCAI grand challenge of mitosis detection.}
		\label{AMIDA-2016-table}
		\begin{tabular}{*{22}{c}}
			\hline\noalign{\smallskip}
			Team & F$_1$-score \\
			\noalign{\smallskip}
			\hline
			\noalign{\smallskip}
			
			Lunit Inc. &0.652\\
			
			IBM Research Zurich and Brazil &0.648\\
			
			Contextvision (SLDESUTO-BOX) &0.616\\
			
			The Chinese University of Hong Kong &0.601\\
			
			Microsoft Research Asia &0.596\\
			
			Radboud UMC &0.541\\
			
			University of Heidelberg &0.481\\
			
			University of South Florida &0.440\\
			
			Pakistan Institute of Engineering and Applied Sciences &0.424\\
			
			University of Warwick &0.396\\
			
			Shiraz University of Technology &0.330\\
			
			Inha University &0.251\\
			
			\noalign{\smallskip}
			\cdashline{1-4}[11pt/3pt]
			\noalign{\smallskip}
			
			CNNCS (on validation set) &0.634\\
			
			\hline
		\end{tabular}
	\end{center}
\end{table}

\begin{figure}[h]
	\centering
	\setlength{\abovecaptionskip}{0pt}
	\setlength{\belowcaptionskip}{0pt}
	\includegraphics[width=13cm]{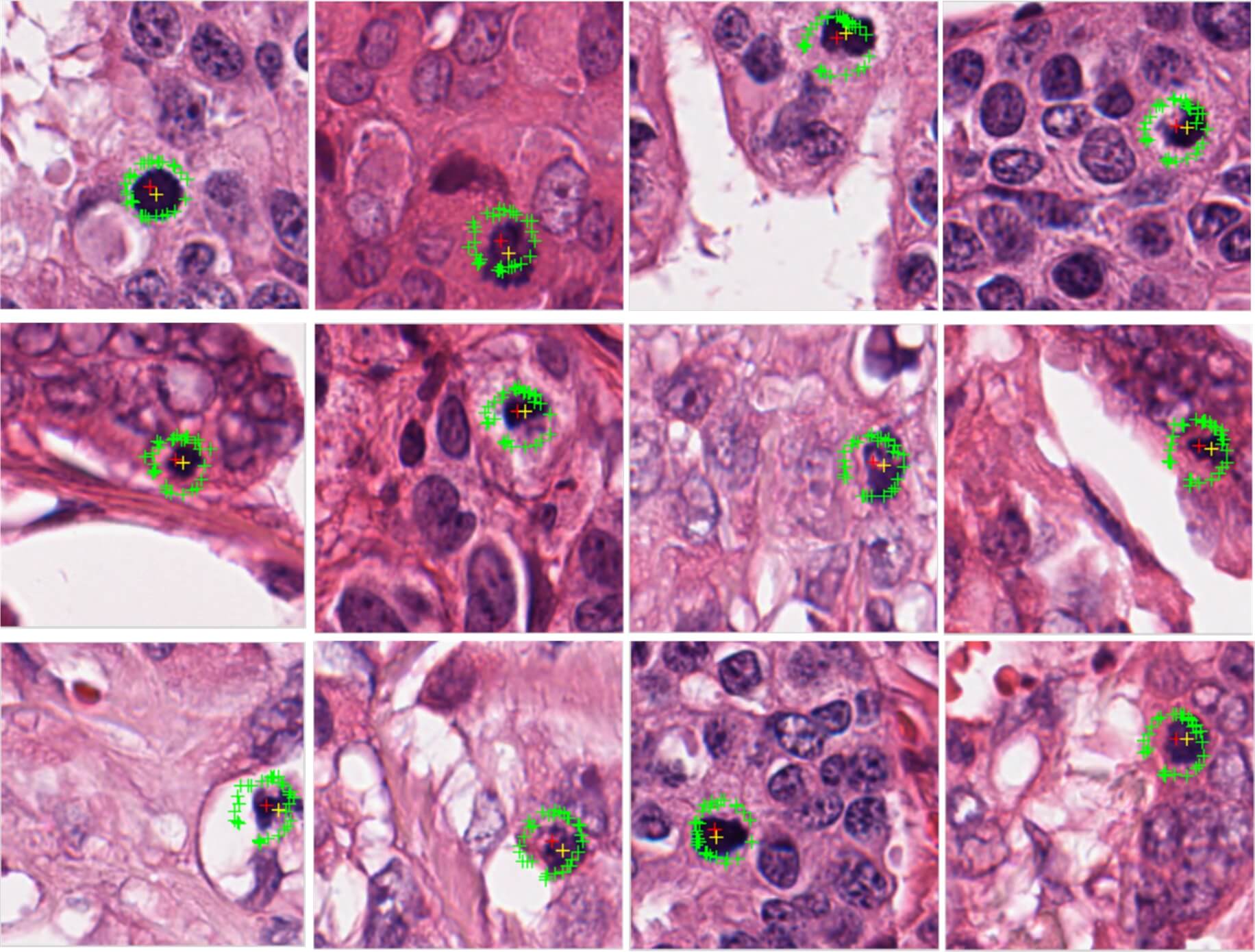}
	\caption{Results on AMIDA-2016 dataset. Yellow cross indicates the ground-truth position of target cells. Green cross indicates cell position predicted by an observation axis. Red cross indicates the final detected cell position, which is the average of all green crosses.}
	\label{200-results}
\end{figure}

We train the proposed CNNCS method using randomly chosen 470 training images and validate on the remaining 117 training images. Additionally, we apply the following ensemble averaging technique to further increase precision and recall values. Originally, we have partitioned every test image into about 100 non-overlapping patches. Instead of starting the partitioning from the top-left corner, now we set the starting point of the first patch from \{offset, offset\}. The offset values are set as 0, 20, 40,..., 160, and 180 (i.e. every 20 pixel) resulting in a total of 10 different settings. Under every offset setting, CNNCS method is run on all the generated patches and provides detection results. Then, we merge detection results from all the offset settings. The merging decision rule is that if there are 6 or more detections within a radius of 9 pixels, then we accept average of these locations as our final detected cell center. Other implementation settings are similar to the settings in the experiment of AMIDA-2013. Finally, we achieved F1-score=0.634 on the validation set (becuase of the lack of publicly available test set), which is the \textbf{third highest} in all the 15 participated teams. Table~\ref{AMIDA-2016-table} provides more details of the contest results. Furthermore, Fig.\ref{200-results} provides twelve examples of our detection results in the AMIDA-2016 grand challenge of mitosis detection.

\section{Conclusion}

This is the first attempt demonstrating that deep convolutional neural network can work in conjunction with compressed sensing-based output encoding schemes toward solving a significant medical image processing task: cell detection and localization from microscopy images. In this work, we made substantial experiments on several mainstream datasets and challenging cell detection contests, where the proposed CNN + CS framework (referred to as CNNCS) achieved very competitive (the highest or at least top-3 in terms of F1-score) results compared to the state-of-the-art methods in cell detection task. In addition, the CNNCS framework has the potential to be trained in End-to-End manner, which is our near future plan and could further boost performance.

\clearpage

\addcontentsline{toc}{section}{References}
\bibliographystyle{plain}
\bibliography{refs}

\begin{thebibliography}{10}

\bibitem{ICPR-2014}
\url{https://mitos-atypia-14.grand-challenge.org/home/}.

\bibitem{AMIDA-2016}
\url{http://tupac.tue-image.nl/}.

\bibitem{LoG-cellSeg}
Yousef Al-Kofahi, Wiem Lassoued, William Lee, and Badrinath Roysam.
\newblock Improved automatic detection and segmentation of cell nuclei in
  histopathology images.
\newblock {\em IEEE Transactions on Biomedical Engineering}, 57:841--852, 2010.

\bibitem{Alpher21}
Carlos Arteta, Victor Lempitsky, J.~Alison Noble, and Andrew Zisserman.
\newblock Learning to detect cells using non-overlapping extremal regions.
\newblock {\em International Conference on Medical Image Computing and
  Computer-Assisted Intervention (MICCAI)}, pages 348--356, 2012.

\bibitem{OMP}
T.~Tony Cai and Lie Wang.
\newblock Orthogonal matching pursuit for sparse signal recovery with noise.
\newblock {\em IEEE Transactions on Information Theory.}, 2011.

\bibitem{bacterial-data}
Alison~Noble Carlos~Arteta, Victor~Lempitsky and Andrew Zisserman.
\newblock Learning to count objects in images.
\newblock {\em Neural Information Processing Systems.}, 2010.

\bibitem{Caruana1997}
Rich Caruana.
\newblock Multitask learning.
\newblock {\em Machine Learning}, 28(1):41--75, Jul 1997.

\bibitem{CasNN}
Hao Chen, Qi~Dou, Xi~Wang, Jing Qin, and Pheng-Ann Heng.
\newblock Mitosis detection in breast cancer histology images via deep cascaded
  networks.
\newblock {\em Proceedings of the Thirtieth Conference on Artificial
  Intelligence (AAAI)}, 2016.

\bibitem{DNN-IDSIA}
Dan~C. Cireşan, Alessandro Giusti, Luca~M. Gambardella, and Jürgen
  Schmidhuber.
\newblock {\em Mitosis Detection in Breast Cancer Histology Images with Deep
  Neural Networks}.
\newblock 2013.

\bibitem{HOG}
Navneet Dalal and Bill Triggs.
\newblock Histograms of oriented gradients for human detection.
\newblock {\em Computer Vision and Pattern Recognition}, 2005.

\bibitem{ECOC}
T.~G. Dietterich and G.~Bakiri.
\newblock Solving multiclass learning problems via error-correcting output
  codes.
\newblock {\em Journal of Artificial Intelligence Research}, pages 263--286,
  1995.

\bibitem{Alpher16}
David~L. Donoho.
\newblock Compressed sensing.
\newblock {\em IEEE Transactions on Information Theory}, 2006.

\bibitem{Alpher17}
Justin~Romberg Emmanuel~Candes.
\newblock Practical signal recovery from random projections.
\newblock {\em IEEE Transactions on Signal Process}, 2005.

\bibitem{Alpher18}
Terence~Tao Emmanuel~Candesy, Justin~Rombergy.
\newblock Robust uncertainty principles: Exact signal reconstruction from
  highly incomplete frequency information.
\newblock {\em IEEE Transactions on Information Theory.}, 2006.

\bibitem{Meijering12}
Meijering Erik.
\newblock Cell segmentation: 50 years down the road [life sciences].
\newblock {\em IEEE Signal Process. Mag.}, 2012.

\bibitem{fast-rcnn}
Ross Girshick.
\newblock Fast r-cnn.
\newblock {\em International Conference on Computer Vision}, 2015.

\bibitem{rcnn}
Ross Girshick, Jeff Donahue, Trevor Darrell, and Jitendra Malik.
\newblock Rich feature hierarchies for accurate object detection and semantic
  segmentation.
\newblock {\em Computer Vision and Pattern Recognition}, 2014.

\bibitem{ResNet}
Kaiming He, Xiangyu Zhang, Shaoqing Ren, and Jian Sun.
\newblock Deep residual learning for image recognition.
\newblock {\em Computer Vision and Pattern Recognition.}, 2015.

\bibitem{CS}
Daniel Hsu, Sham~M. Kakade, John Langford, and Tong Zhang.
\newblock Multi-label prediction via compressed sensing.
\newblock {\em arXiv:0902.1284v2 [cs.LG]}, 2009.

\bibitem{IPAL}
Humayun Irshad.
\newblock Automated mitosis detection in histopathology using morphological and
  multi-channel statistics features.
\newblock {\em Journal of Pathology Informatics}, 2013.

\bibitem{output-space-thesis}
Arnaud Joly.
\newblock Exploiting random projections and sparsity with random forests and
  gradient boosting methods.
\newblock {\em arXiv:1704.08067}, 2016.

\bibitem{Nuclei-data}
Y.W Tsang I.A. Cree D.R.J.~Snead K.~Sirinukunwattana, S.E.A.~Raza and N.M.
  Rajpoot.
\newblock Locality sensitive deep learning for detection and classification of
  nuclei in routine colon cancer histology images.
\newblock {\em IEEE Transactions on Medical Imaging.}, 2016.

\bibitem{Bayesian-CS}
Ashish Kapoor, Raajay Viswanathan, and Prateek Jain.
\newblock Multilabel classification using bayesian compressed sensing.
\newblock {\em Advances in Neural Information Processing Systems}, 2012.

\bibitem{LoG}
Hui Kong, Hatice~Cinar Akakin, and Sanjay~E. Sarma.
\newblock A generalized laplacian of gaussian filter for blob detection and its
  applications.
\newblock {\em IEEE Transactions on Cybernetics}, 43:1719--1733, 2013.

\bibitem{ImageNet}
Alex Krizhevsky, Ilya Sutskever, and Geoffrey~E. Hinton.
\newblock Imagenet classiﬁcation with deep convolutional neural networks.
\newblock {\em Neural Information Processing Systems}, 2012.

\bibitem{LeCun-nature}
Yann LeCun, Yoshua Bengio, and Geoffrey Hinton.
\newblock Deep learning.
\newblock {\em Nature}, 521:436--444, 2015.

\bibitem{sift}
David~G. Lowe.
\newblock Object recognition from local scale-invariant features.
\newblock In {\em Proceedings of the International Conference on Computer
  Vision-Volume 2 - Volume 2}, ICCV '99, pages 1150--, Washington, DC, USA,
  1999. IEEE Computer Society.

\bibitem{NEC}
Christopher~D. Malon and Eric Cosatto.
\newblock Classification of mitotic figures with convolutional neural networks
  and seeded blob features.
\newblock {\em Journal of Pathology Informatics}, 2013.

\bibitem{LBP}
T.~Ojala, M.~Pietikäinen, and D.~Harwood.
\newblock A comparative study of texture measures with classification based on
  feature distributions.
\newblock {\em Pattern Recognition}, 29:51--59, 1996.

\bibitem{redmon2016yolo9000}
Joseph Redmon and Ali Farhadi.
\newblock Yolo9000: Better, faster, stronger.
\newblock {\em arXiv preprint arXiv:1612.08242}, 2016.

\bibitem{faster-rcnn}
Shaoqing Ren, Kaiming He, Ross Girshick, and Jian Sun.
\newblock Faster {R-CNN}: Towards real-time object detection with region
  proposal networks.
\newblock {\em Advances in Neural Information Processing Systems}, 2015.

\bibitem{ICPR-2012}
Ludovic Roux, Daniel Racoceanu, N.~Lom{\'e}nie, Maria~S. Kulikova, Humayun
  Irshad, J.~Klossa, Fr{\'e}d{\'e}rique Capron, Catherine Genestie, Gilles~Le
  Naour, and Metin~N. Gurcan.
\newblock Mitosis detection in breast cancer histological images, an icpr 2012
  contest.
\newblock {\em Journal of Pathology Informatics}, 4, 05/2013 2013.

\bibitem{AggNet16}
Albarqouni Shadi, Baur Christoph, Achilles Felix, Belagiannis Vasileios,
  Demirci Stefanie, and Navab Nassir.
\newblock Aggnet: Deep learning from crowds for mitosis detection in breast
  cancer histology images.
\newblock {\em IEEE Transactions on Medical Imaging.}, 2016.

\bibitem{FCN}
Evan Shelhamer, Jonathan Long, and Trevor Darrell.
\newblock Fully convolutional models for semantic segmentation.
\newblock {\em The IEEE Transactions on Pattern Analysis and Machine
  Intelligence}, 2016.

\bibitem{DAL}
Ryota Tomioka, Taiji Suzuki, and Masashi Sugiyama.
\newblock Super-linear convergence of dual augmented lagrangian algorithm for
  sparsity regularized estimation.
\newblock {\em The Journal of Machine Learning Research}, 2011.

\bibitem{RAkEL}
Grigorios Tsoumakas, Ioannis Katakis, and Ioannis Vlahavas.
\newblock Random k-labelsets for multi-label classification.
\newblock {\em IEEE Transactions on Knowledge and Data Engineering}, pages
  1079--1089, 2011.

\bibitem{AMIDA-2013}
Mitko Veta, Paul~J van Diest, Stefan~M Willems, Haibo Wang, Anant Madabhushi,
  Angel Cruz-Roa, Fabio Gonzalez, Anders B~L Larsen, Jacob~S Vestergaard,
  Anders~B Dahl, Dan~C Cire{\c s}an, J{\"u}rgen Schmidhuber, Alessandro Giusti,
  Luca~M Gambardella, F~Boray Tek, Thomas Walter, Ching-Wei Wang, Satoshi
  Kondo, Bogdan~J Matuszewski, Frederic Precioso, Violet Snell, Josef Kittler,
  Teofilo~E de~Campos, Adnan~M Khan, Nasir~M Rajpoot, Evdokia Arkoumani,
  Miangela~M Lacle, Max~A Viergever, and Josien P~W Pluim.
\newblock Assessment of algorithms for mitosis detection in breast cancer
  histopathology images.
\newblock {\em Medical image analysis}, 20:237--48, 02/2015 2015.

\bibitem{Alpher25}
Weidi Xie, J.~Alison Noble, and Andrew Zisserman.
\newblock Microscopy cell counting with fully convolutional regression
  networks.
\newblock {\em Deep Learning Workshop in MICCAI}, 2015.

\bibitem{our-ECCV}
Yao Xue, Nilanjan Ray, Judith Hugh, and Gilbert Bigras.
\newblock Cell counting by regression using convolutional neural network.
\newblock {\em ECCV 2016 workshop on BioImage Computing}, 2016.

\end{thebibliography}

\end{document}